\documentclass[letterpaper, 10 pt, conference]{ieeeconf}
\IEEEoverridecommandlockouts
\usepackage{makecell}
\usepackage{amsmath,amssymb}
\usepackage{booktabs}
\usepackage{stfloats}
\usepackage{float}
\usepackage{array,tabularx}
\usepackage{cite}
\usepackage{url}
\usepackage{microtype}
\usepackage{pifont}
\usepackage{graphicx}
\usepackage[table]{xcolor}
\usepackage[hidelinks]{hyperref}
\newcommand{\cmark}{\textcolor{green!55!black}{\ding{51}}}
\newcommand{\xmark}{\textcolor{red!80!black}{\ding{55}}}

\title{\LARGE \bf CAPER: PartGround and PlanRefine for Reliable Manual-Guided Robotic Assembly Planning}

\author{\normalsize
Hanzhao Xi$^{1}$, Zhonghao Li$^{2}$, Suxuan Liu$^{2}$,
Shanghan Zhou$^{2}$, Zesheng Chen$^{2}$, Qiufeng Wang$^{3}$,
Yaran Chen$^{*}$\thanks{\raggedright $^{*}$Corresponding author: Yaran Chen (email: \href{mailto:yaran.chen@xjtlu.edu.cn}{\texttt{yaran.chen@xjtlu.edu.cn}}).}\\[0.2em]
\small Xi'an Jiaotong-Liverpool University
}
\hypersetup{
  pdftitle={CAPER: PartGround and PlanRefine for Reliable Manual-Guided Robotic Assembly Planning},
  pdfauthor={Hanzhao Xi, Zhonghao Li, Suxuan Liu, Shanghan Zhou, Zesheng Chen, Qiufeng Wang, Yaran Chen}
}

\IEEEaftertitletext{\vspace{-\baselineskip}}

\begin{document}
\maketitle
\begin{abstract}
  Vision-Language-Models (VLMs) used in robot task planning to derive assembly procedures from existing manuals may correctly interpret individual assembly actions yet confuse exact part identities, predict unsupported Part IDs, or produce inconsistent parts across assembly steps. Thus, we propose CAPER (Capacity-Aware Part-grounding and Evidence Refinement), a manual-guided robotic assembly framework that enables VLMs to generate assembly procedures with two modules—PartGround and PlanRefine. PartGround aligns step-level manual evidence with product-specific inventory and part-reference information to identify exact Part IDs and quantities, transforming free-form hardware prediction into evidence-guided assignments represented in a shared Assembly Intermediate Representation (Assembly IR). PlanRefine audits PartGround assignments against finite inventory by removing the lowest-confidence assignments. Uncertain identity cases remain marked for review in the Assembly IR. We evaluate CAPER on the IKEA Assembly in the Wild (IAW) dataset. Experiment results show that CAPER reaches a Finite-inventory Consistency Rate of 83.33\% and produces zero observed out-of-inventory Part-ID occurrences.  At the plan level, PlanRefine removes all observed inventory-overuse pairs (9$\rightarrow$0) and excess units (51$\rightarrow$0). Overall, the results indicate that PartGround and PlanRefine improve part reliability, while also enhancing finite-inventory consistency and local prediction accuracy.
    \par\noindent Project page and video: \href{https://xhzzzz123.github.io/CAPER/}{xhzzzz123.github.io/CAPER/}.
\end{abstract}

 \begin{table*}[t]
    \centering
    \caption{Functional coverage of assembly planning methods.}
    \label{tab:relatedcomparison}
    \scriptsize
    \setlength{\tabcolsep}{2.7pt}
    \renewcommand{\arraystretch}{1.18}
    \begin{tabularx}{\textwidth}{>{\raggedright\arraybackslash}p{2.45cm} *{7}{>{\centering\arraybackslash}X}}
    \toprule
    Method & Assembly Manual Input & Structured Assembly Plan & Exact Part-ID Matching & Part Quantities & Finite-inventory Consistency & Review / Abstention Support & Reviewable \\
    \midrule
    Sera et al.~\cite{sera2021graphicalmanual} & \cmark & \cmark & \xmark & \xmark & \xmark & \xmark & \xmark \\
    SayCan~\cite{ichter2022saycan} & \xmark & \xmark & \xmark & \xmark & \xmark & \xmark & \xmark \\
    ProgPrompt~\cite{singh2023progprompt} & \xmark & \cmark & \xmark & \xmark & \xmark & \xmark & \xmark \\
    KnowNo~\cite{ren2023knowno} & \xmark & \xmark & \xmark & \xmark & \xmark & \cmark & \xmark \\
    Manual2Skill~\cite{tie2025manual2skill} & \cmark & \cmark & \xmark & \xmark & \xmark & \xmark & \xmark \\
    VLM-TAMP~\cite{yang2025vlmtamp} & \xmark & \cmark & \xmark & \xmark & \xmark & \xmark & \xmark \\
    Manual2Skill++~\cite{tie2026manual2skillpp} & \cmark & \cmark & \xmark & \cmark & \xmark & \xmark & \xmark \\
    \textbf{CAPER (Ours)} & \cmark & \cmark & \cmark & \cmark & \cmark & \cmark & \cmark \\
    \bottomrule
    \end{tabularx}
    \\[2.5mm]
    \parbox{0.98\textwidth}{\scriptsize \raggedright
    Note: Relative to the other rows, CAPER adds \textbf{Exact Part-ID Matching}, \textbf{Finite-inventory Consistency}, and \textbf{Reviewable}, which are unmarked for every compared method. It also includes \textbf{Part Quantities} (shared only with Manual2Skill++) and \textbf{Review / Abstention Support} (shared only with KnowNo), making it the only method marked for all seven functions. Within CAPER, \textbf{Exact Part-ID Matching} and \textbf{Part Quantities} are produced by PartGround, \textbf{Finite-inventory Consistency} is produced by PlanRefine, and the two review-related columns are jointly realized through the shared Assembly IR. This added coverage is important for strengthening robot-arm assembly: the plan then specifies exact physical parts and remains consistent with finite inventory, rather than stopping at step semantics.\par}
    \end{table*}
    
    \section{Introduction}
    Enabling robots to derive assembly procedures from human-readable manuals provides a practical path toward flexible manufacturing. Rather than programming each product individually, a robot can use existing illustrations, annotations, and part references to infer the required operations and components. Recent vision-language models (VLMs) make this increasingly feasible by combining high-level task knowledge with estimates of skill feasibility~\cite{ichter2022saycan}, or by generating structured plans from available actions, objects, and example programs~\cite{singh2023progprompt}.


    Recent studies have explored this problem from complementary perspectives. Manual2Skill extracts hierarchical assembly graphs from instructional images and couples them with pose estimation and robot execution ~\cite{tie2025manual2skill}, while Manual2Skill++ further incorporates connector-level information, including connector type, specification, quantity, and spatial relations, for more detailed assembly reasoning ~\cite{tie2026manual2skillpp}. Related planning methods such as SayCan ~\cite{ichter2022saycan}, ProgPrompt \cite{singh2023progprompt}, and VLM-TAMP \cite{yang2025vlmtamp} introduce affordance, structural, or task-and-motion constraints, while KnowNo \cite{ren2023knowno} considers uncertainty-aware intervention. To better position the reliability problem studied in this work, Table\ref{tab:relatedcomparison} summarizes the  representative methods in terms of manual input, structured planning, exact Part-ID matching, part quantities, finite-inventory consistency, reviewability, which is intended to characterize the information and constraints handled by each method. Yet semantic understanding alone cannot ensure correct part grounding or sequence-wide feasibility.

    At the step level, a key difficulty is  \textbf{exact part-identity grounding}. A model may correctly recognize that a step requires four screws while still failing to determine which specific screws in the product inventory should be used. Assembly manuals often contain simplified drawings, small hardware illustrations, visually similar components, and sparse or distributed annotations, making this correspondence difficult to establish. Manual2Skill++ \cite{tie2026manual2skillpp}, for example, explicitly represents connector quantities and specifications, but this is distinct from grounding each assembly step to the exact product-specific Part ID. Consequently, a model may generate a Part ID that is absent from the inventory or select a valid but incorrect candidate. This distinction is critical for robotic assembly, where the downstream system must retrieve and manipulate a specific physical component rather than merely recognize its semantic category.

    At the plan level, locally plausible assignments may still become inconsistent when combined across the complete assembly sequence. Multiple steps may collectively consume more instances of a Part ID than are available, while ambiguous evidence may lead to conflicting assignments across steps. Existing planning methods address related forms of reliability, including skill affordances \cite{ichter2022saycan}, structured task constraints \cite{singh2023progprompt}, task-and-motion feasibility 
\cite{yang2025vlmtamp}, and uncertainty-aware intervention \cite{ren2023knowno}. However,\textbf{ among the representative methods summarized in Table \ref{tab:relatedcomparison}, exact Part-ID grounding, part quantities, and finite-inventory consistency are not jointly modeled as a unified material-level constraint over the complete assembly plan}. Reliable assembly planning therefore requires not only semantically plausible local steps but also physically grounded assignments that remain consistent with the available inventory throughout the procedure.

    These two reliability gaps require different but complementary forms of reasoning. Exact part identification requires aligning step-level manual evidence with product-specific inventory and part-reference information, whereas plan validation requires examining how the grounded parts are accumulated and reused across the full sequence. A structured plan alone cannot guarantee either property: a well-formed prediction may still select the wrong physical part, while an assignment restricted to legal Part IDs may still choose the wrong legal candidate. Likewise, an assignment that appears reasonable locally may violate finite-inventory constraints when aggregated across multiple steps. This motivates a framework that connects evidence-based physical grounding with plan-level consistency checking through a shared structured representation.

    To this end, we propose \textbf{CAPER (Capacity-Aware Part-grounding and Evidence Refinement)}, a manual-guided assembly planning framework composed of two complementary modules: \textbf{PartGround} and \textbf{PlanRefine}. CAPER represents each assembly step using a shared \textbf{Assembly Intermediate Representation (Assembly IR)} that records the operation, exact Part IDs, quantities, and supporting manual evidence. PartGround addresses part-identity hallucinations by linking step-level visual evidence with product-specific inventory references, transforming free-form hardware prediction into evidence-guided physical grounding. PlanRefine addresses plan inconsistency by auditing the grounded assignments against finite inventory: when predicted consumption exceeds available quantity, it removes the lowest-confidence assignments until finite-inventory consistency is restored. In this way, PartGround focuses on the question of which physical part belongs to each step, while PlanRefine focuses on whether the resulting assignments satisfy finite-inventory consistency. Together, the two modules connect semantic manual understanding with exact physical part grounding and plan-level inventory validation, rather than treating a single VLM prediction as the final assembly decision.
    
    We evaluate CAPER on the \emph{IKEA Assembly in the Wild (IAW)} dataset~\cite{zhang2023iaw} through model adaptation, Part-ID grounding, and failure-aware end-to-end evaluation. The experiments show that additional model adaptation alone does not consistently resolve exact physical grounding, motivating explicit reliability mechanisms outside the VLM. CAPER produces zero observed out-of-inventory Part-ID occurrences and reaches an 83.33\% Finite-inventory Consistency Rate. At the plan level, PlanRefine removes all observed cross-step inventory-overuse pairs (9$\rightarrow$0) and excess units (51$\rightarrow$0). Part-ID F1 changes only slightly from 55.24\% before PlanRefine to 55.77\% after refinement, showing that the primary role of PlanRefine is finite-inventory consistency rather than local identity accuracy. Taken together, these results show that legal part selection, correct physical grounding, and finite-inventory consistency are related but distinct reliability requirements. They also indicate that the most difficult remaining problem is not simply restricting outputs to legal IDs, but selecting the correct legal physical identity when the manual evidence is ambiguous.
    
    Our contributions are threefold:
\vspace{-0.35em}
\begin{itemize}
    \setlength{\topsep}{0pt}
    \setlength{\partopsep}{0pt}
    \setlength{\itemsep}{1pt}
    \setlength{\parsep}{0pt}
    \item \textbf{A reliability-oriented framework for manual-guided assembly planning.}
    We identify two key gaps between semantic assembly understanding and reliable physical execution: exact product-specific Part-ID grounding and cross-step finite-inventory consistency. To address them, we propose CAPER, a unified framework that connects structured manual understanding with physical part grounding and plan-level inventory validation through a shared Assembly IR.

    \item \textbf{Evidence-guided exact Part-ID grounding.}
    We introduce PartGround to resolve part-identity ambiguity by aligning step-level manual evidence with product-specific inventory and part-reference information, and retrieving additional evidence when necessary. This converts free-form hardware prediction into evidence-supported assignments of exact Part IDs and quantities.

    \item \textbf{Finite-inventory-aware plan refinement.}
    We introduce PlanRefine to audit grounded Part-ID assignments across the complete assembly sequence and repair inventory conflicts by removing the lowest-confidence assignments when predicted consumption exceeds available quantities, while preserving unresolved cases for review.
\end{itemize}
    
    \section{Related Work}
    \subsection{Assembly Planning from Human Instructions}
    Early work showed that graphical manuals can be converted into structured robot assembly plans. Sera et al. build an Assembly Task Sequence Graph from a graphical instruction manual, infer missing information across the image sequence, and check total part counts before robot execution~\cite{sera2021graphicalmanual}. Manual2Skill uses a VLM to extract hierarchical assembly graphs and couples them with pose estimation, motion planning, and real IKEA furniture assembly~\cite{tie2025manual2skill}. Manual2Skill++ extends this line with connector-aware graphs that represent connector type, spatial relation and so on~\cite{tie2026manual2skillpp}. LLM-as-BT-Planner studies another structured representation, generating behavior trees for robotic assembly tasks~\cite{ao2025btplanner}. These works establish the value of structured representations for instruction-guided assembly. CAPER focuses on other questions: whether each step is linked to the exact product-specific hardware identity and whether those assignments remain consistent with the inventory of final plan.

    \subsection{Grounded Robot Planning}
    Language- and vision-based robot planning often separates semantic reasoning from feasibility checks. SayCan combines language-model preferences with robot skill affordances~\cite{ichter2022saycan}. ProgPrompt constrains generated plans with program-like actions, objects, and preconditions~\cite{singh2023progprompt}. VLM-TAMP uses VLM-generated subgoals to guide task-and-motion planning and replans when a subgoal cannot be refined into a feasible motion plan~\cite{yang2025vlmtamp}. These methods address action or motion feasibility in the robot environment. CAPER operates at the earlier layers: it checks whether the selected physical hardware is supported by the manual and whether the complete set of Part-ID assignments respects finite inventory.
    
    \subsection{Uncertainty and Failure Awareness}
    KnowNo uses conformal prediction to identify uncertain language-model plans and request human help when needed~\cite{ren2023knowno}. Inner Monologue\cite{huang2023inner} and SymSkill\cite{shao2026symskill} also use explicit feedback or recovery mechanisms to improve long-horizon robot behavior. CAPER's review is tied to evidence quality, unresolved Part-ID assignments, upstream gate failures, and inventory conflicts. This distinction is important because a grounding-quality state and a probability of task success are different quantities.
    
    Table~\ref{tab:relatedcomparison} summarizes the problem coverage of representative methods. The table is a capability comparison since the methods use different datasets, tasks, and evaluation protocols.

 \begin{figure*}[!t]
    \centering
    \includegraphics[width=\textwidth]{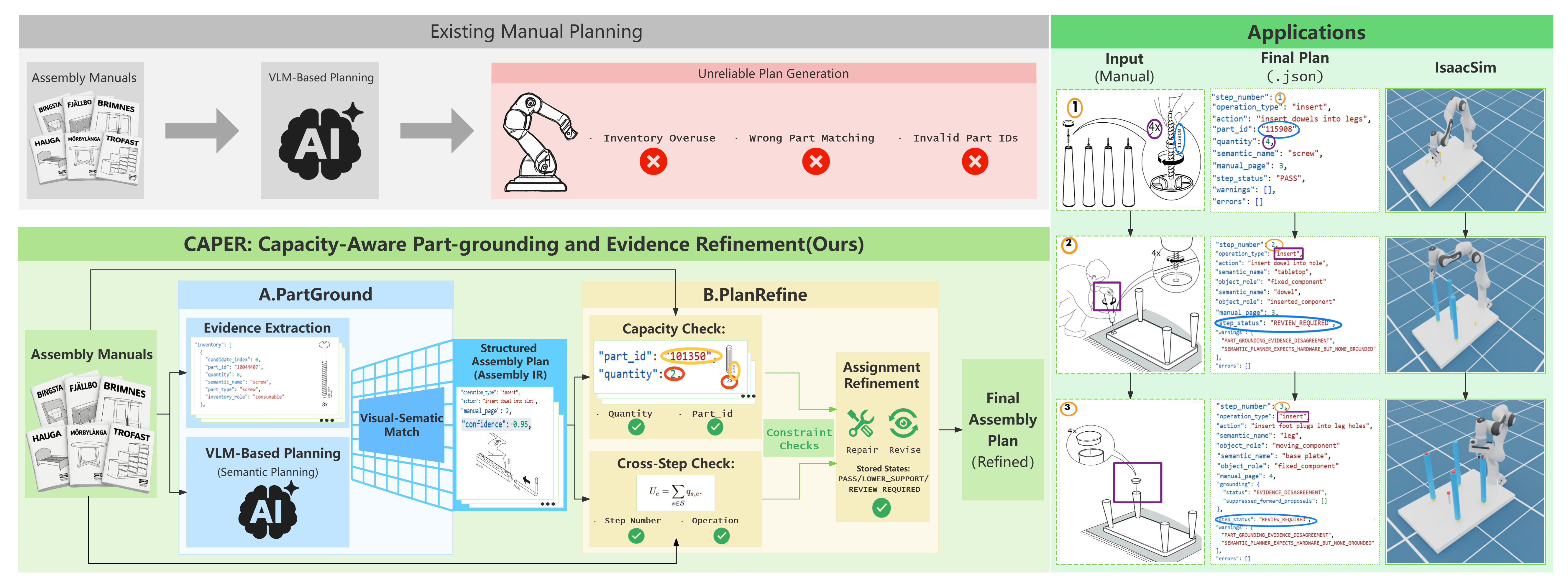}
      \caption[Overview of CAPER]{%
    {
    \noindent\textbf{Overview of CAPER: Capacity-Aware Part-grounding and Evidence Refinement for manual-guided assembly planning.}\par}
    \textbf{Top left:} Direct VLM-based planning may produce several problems.
    \textbf{Bottom left:} \textbf{(A) PartGround} combines  manual evidence, inventory references, and semantic predictions to ground Part IDs and quantities in Assembly IR.
    \textbf{(B) PlanRefine} audits the grounded assignments, removes low-confidence assignments, and retains unresolved cases for review.
    \textbf{Right:} Example manual steps, structured JSON outputs, and corresponding assembly scenes in IsaacSim.
}
    \label{fig:caper_overview}
\end{figure*}

    \section{CAPER Framework}
\label{sec:caper_framework}

\subsection{Overall Framework}
\label{subsec:overall_framework}

As illustrated in Fig.\ref{fig:caper_overview}, CAPER is designed to address
two reliability problems in manual-guided robotic assembly planning:
exact physical part identification and finite-inventory consistency.
Although a vision-language model (VLM) can recover the semantic meaning
of an assembly instruction, direct generation may still produce
unsupported Part IDs, confuse visually similar components, or generate
part assignments that become inconsistent across multiple steps. CAPER
therefore uses the VLM only as a semantic planner and delegates these two
problems to two dedicated modules: \textbf{PartGround} and
\textbf{PlanRefine}. PartGround grounds semantic hardware descriptions to
product-specific Part IDs, while PlanRefine checks whether the resulting
assignments remain consistent with the finite inventory over the complete
assembly sequence.

CAPER represents the intermediate and final results using a shared
\textbf{Assembly Intermediate Representation (Assembly IR)}. For each
assembly step, the Assembly IR records the predicted operation, action
description, grounded Part IDs, quantities, supporting evidence, warnings,
and grounding-quality information. This representation allows semantic
understanding, physical grounding, and plan-level consistency checking to
remain explicitly connected. The main design of CAPER is therefore to
separate semantic reasoning from exact Part-ID determination and
cross-step inventory validation, rather than relying on a single VLM
prediction as the final assembly decision. Fig.~\ref{fig:caper_overview}
illustrates this organization and shows how the structured output is
connected to downstream robotic assembly.

\subsection{PartGround: Evidence-Guided Part-ID Grounding}
\label{subsec:partground}

PartGround addresses the question of which exact physical part should be
used in each assembly step. A semantic prediction such as ``screw'' or
``connector'' is insufficient for robotic execution when several
product-specific candidates have similar appearance or function.
PartGround therefore reformulates Part-ID prediction from unrestricted
generation into an inventory-constrained grounding problem. For each
assembly step $s$, it receives the localized manual evidence $E_s$, the
semantic prediction $M_s$, the product-specific candidate inventory
$\mathcal{C}$, and visual references of the candidate parts. The candidate
inventory is represented as
\begin{equation}
\mathcal{C}
=
\left\{
c_i=(id_i,Q_i,r_i,v_i)
\right\}_{i=1}^{N},
\label{eq:candidate_inventory}
\end{equation}
where $id_i$ denotes the product-specific Part ID, $Q_i$ its available
quantity, $r_i$ its semantic or reference information, and $v_i$ its
visual reference.

The evidence required by PartGround is extracted directly from the
assembly manual. Instruction, inventory, and reference regions are
identified and associated with individual assembly steps, while candidate
Part IDs, quantities, and reference locations are extracted to construct
the candidate inventory. When visual extraction from a searchable PDF is
unreliable, numeric text anchors are used as deterministic recovery cues.
Qwen3-VL-4B-Instruct~\cite{bai2025qwen3vl} then predicts the operation,
action description, and semantic hardware requirements for each localized
step. Numeric Part IDs predicted by the VLM are not accepted as final
physical identities.

PartGround combines the semantic prediction with candidate references,
manual evidence, and explicit hardware or quantity annotations when
available. The selected physical identity is constrained by
\begin{equation}
\hat{c}_s \in \mathcal{C},
\label{eq:legal_candidate}
\end{equation}
so the final Part ID must correspond to a candidate extracted from the
product-specific inventory. Agreement between manual callouts and
candidate-reference evidence provides stronger support, whereas conflicting
or incomplete evidence is retained explicitly rather than silently
overwritten. The output of PartGround is
\begin{equation}
A_s^{\mathrm{PG}}
=
\left(
\hat{c}_s,
\hat{q}_s,
\gamma_s,
\mathcal{E}_s
\right),
\label{eq:partground_output}
\end{equation}
where $\hat{c}_s$ and $\hat{q}_s$ denote the grounded Part ID and quantity,
$\gamma_s$ denotes grounding support, and $\mathcal{E}_s$ records the
supporting evidence.

The contribution of PartGround is therefore more than restricting outputs
to legal identifiers. A legal Part ID can still correspond to the wrong
physical component. By linking semantic hardware requirements with
product-specific visual and inventory evidence, PartGround produces
explicit and inspectable physical assignments that can be directly used
by both PlanRefine and downstream robotic part retrieval and manipulation.

\subsection{PlanRefine: Finite-Inventory Consistency Refinement}
\label{subsec:planrefine}

PartGround produces step-level physical assignments, but individually
plausible assignments do not guarantee that the complete assembly plan is
consistent with the available inventory. PlanRefine therefore operates on
the complete set of grounded assignments and checks whether repeated use of
each Part ID exceeds its available quantity. It does not perform additional
Part-ID inference and does not search for alternative candidates.

For each Part ID $c$, let $Q_c$ denote its available quantity and
$q_{s,c}$ the quantity assigned to step $s$. The total predicted
consumption is
\begin{equation}
U_c =
\sum_{s \in S} q_{s,c}.
\label{eq:total_consumption}
\end{equation}
If $U_c \leq Q_c$, all assignments associated with $c$ are retained. If
$U_c > Q_c$, PlanRefine orders the corresponding assignments by increasing
grounding support, with later steps removed first when support values are
equal. Let $\pi_c(1),\ldots,\pi_c(n_c)$ denote this ordering. PlanRefine
removes the smallest number of assignments required to restore the
inventory constraint:
\begin{equation}
k_c^{*}
=
\min
\left\{
k:
U_c-
\sum_{i=1}^{k}
q_{\pi_c(i),c}
\leq Q_c
\right\}.
\label{eq:minimal_repair}
\end{equation}

This refinement follows a minimal-intervention principle: only assignments
associated with an overused Part ID are modified, and the process stops
once the available capacity is restored. No new Part ID is invented, no
replacement candidate is searched, and no combinatorial reassignment is
performed. The resulting plan is re-audited against the same finite
inventory before being emitted.

The refined assignments are stored in the final Assembly IR together with
the semantic operation, action description, Part IDs, quantities,
supporting evidence, and warnings. CAPER additionally retains three
grounding-quality states: \texttt{PASS}, \texttt{LOWER SUPPORT}, and
\texttt{REVIEW REQUIRED}, allowing uncertain assignments to remain
explicit rather than being converted into confident predictions.
PartGround and PlanRefine therefore provide complementary functions:
PartGround determines which physical component belongs to an individual
step, while PlanRefine verifies that these assignments remain jointly
consistent across the complete assembly sequence.

\subsection{Final Assembly Plan}
\label{subsec:final_plan}

The complete CAPER framework shown in Fig.~\ref{fig:caper_overview} connects
semantic instruction understanding with physically grounded and
finite-inventory-consistent assembly planning. The semantic planner
provides the operation and hardware requirement, PartGround resolves the
corresponding product-specific Part IDs and quantities, and PlanRefine
checks the resulting assignments over the complete assembly sequence. The
final Assembly IR therefore combines semantic actions, exact physical
parts, quantities, supporting evidence, and consistency information in a
single structured representation.

As illustrated on the right side of Fig.\ref{fig:caper_overview}, this
structured output provides an explicit interface between manual
understanding and downstream robotic execution. Exact Part IDs identify
the physical components to be localized and retrieved, quantities specify
the required number of parts, and operation information defines the
intended assembly action. CAPER thus converts open-ended VLM reasoning into
an evidence-grounded, product-specific, and finite-inventory-consistent
assembly plan that can support subsequent perception, grasping,
manipulation, and robotic furniture assembly.

\section{Experiments}
\label{sec:caper-experimental}

\subsection{Overview}

We evaluate CAPER on the IKEA Assembly in the Wild (IAW) dataset~\cite{zhang2023iaw} to examine whether evidence-guided part grounding and finite-inventory validation improve manual-guided assembly planning. The experiments compare CAPER with direct VLM generation and a same-backbone Manual2Skill baseline~\cite{tie2025manual2skill}, followed by a cumulative component ablation. We distinguish local prediction quality from plan-level inventory consistency and retain pipeline non-completions in the evaluation denominator.

\subsection{Experimental Setup}
\label{sec:caper-experimental-setup}

\subsubsection{Data}

We evaluate CAPER on the IAW test set~\cite{zhang2023iaw},
which contains 15 furniture instances and 297 ground-truth
assembly steps. None of these furniture instances was used
in our model training or adaptation, so all 15 instances
are previously unseen by the models in our experimental
setting. The pipeline produces outputs for 10 instances,
covering 168 ground-truth steps. Three steps are omitted
during localization, leaving 165 aligned steps. The remaining
5 instances and all missing steps are retained in the full
15-instance / 297-step evaluation denominator and counted
as failures. This strict, failure-aware protocol evaluates
performance across the entire test set rather than only on
instances for which the pipeline completes.

\subsubsection{Baselines}

We compare CAPER against three controlled direct-generation baselines based on the same Qwen3-VL-4B-Instruct~\cite{bai2025qwen3vl}: the original Base model, Supervised Fine-Tuning (SFT), and Direct Preference Optimization (DPO) \cite{rafailov2023dpo}.. These baselines use the same localized visual evidence, prompt and output schema, deterministic decoding protocol, and evaluation procedure, with only the model checkpoint changed; they directly generate the final structured report without PartGround or PlanRefine. We also include Manual2Skill~\cite{tie2025manual2skill} as a same-backbone controlled baseline while the remaining Manual2Skill planning procedure is kept unchanged and no CAPER-specific inventory grounding, PartGround, PlanRefine, or evaluator feedback is provided during inference. CAPER uses the same 4B backbone as a semantic planner and adds PartGround for exact Part-ID grounding and PlanRefine for finite-inventory consistency. Predictions from all methods are frozen before evaluation and scored using the same structured-report evaluation protocol. The numerical comparison with Manual2Skill therefore concerns this controlled planning baseline.

We conduct a cumulative component ablation using the complete CAPER framework, CAPER without PlanRefine, and the semantic planner without both PartGround and PlanRefine. The first comparison isolates the effect of finite-inventory repair, whereas the second additionally removes explicit Part-ID grounding. The semantic-planner-only configuration predicts operations and hardware mentions and is distinct from the Base direct-generation baseline. Because PlanRefine consumes grounded Part IDs, retaining PlanRefine while removing PartGround is not a well-defined independent ablation.

\subsubsection{Metrics}

We report Structured Report Precision, Recall, and F1, where the report is evaluated over operation facts and hardware facts defined by step-level Part ID and quantity. We additionally report Step Report Accuracy, Part-ID F1, and Hardware Exact Accuracy to measure local prediction quality. These metrics are evaluated under the full 15-instance / 297-step failure-aware protocol.

To evaluate whether the Part-ID assignments produced by PartGround and refined by PlanRefine remain consistent with the available parts across the complete assembly plan, we report Finite-inventory Consistency Rate and Complete-output Finite-inventory Consistency Rate on the common 12-instance subset with frozen and verifiable inventory references. Three instances are excluded from these two rates because they lack verifiable frozen inventory references. We also report Out-of-inventory Part-ID Occurrences to measure predictions unsupported by the supplied inventory. This setting allows us to evaluate local prediction accuracy on the full IAW test set~\cite{zhang2023iaw} benchmark while separately examining whether the final assembly plans satisfy finite-inventory consistency.

 \begin{table*}[!t]
    \centering
    \caption{Assembly Planning Performance comparison.}
    \label{tab:overall}
    \scriptsize
    \setlength{\tabcolsep}{3.0pt}
    \renewcommand{\arraystretch}{1.16}
    \begin{tabularx}{\textwidth}{>{\raggedright\arraybackslash}p{5.45cm} *{6}{>{\centering\arraybackslash}X}}
    \toprule
   Method & \makecell[t]{Precision\\(\%) $\uparrow$} & \makecell[t]{Recall\\(\%) $\uparrow$} & \makecell[t]{F1 Score\\(\%) $\uparrow$} & \makecell[t]{Finite-inventory\\ Consistency \\Rate (\%) $\uparrow$} & \makecell[t]{Complete-output\\Finite-inventory\\ Consistency\\Rate (\%) $\uparrow$} & \makecell[t]{Out-of-inventory\\Part-ID\\Occurrences $\downarrow$} \\
    \midrule
    Qwen3-VL-4B-Instruct (Base) & 60.07 & 30.96 & 40.86 & 50.00 & 25.00 & 1 \\
    Qwen3-VL-4B-Instruct + Supervised Fine-Tuning & 62.20 & 30.38 & 40.83 & 66.67 & 41.67 & 2 \\
    Qwen3-VL-4B-Instruct + Direct Preference Optimization & 60.30 & 30.96 & 40.91 & 50.00 & 33.33 & 2 \\
    Manual2Skill~\cite{tie2025manual2skill} & 27.08 & \textbf{33.85} & 30.09 & 41.67 & 33.33 & 118 \\
        \rowcolor{green!15}
    \textbf{CAPER (Ours)} & \textbf{62.99} & 30.77 & \textbf{41.34} & \textbf{83.33} & \textbf{58.33} & \textbf{0} \\
    \bottomrule
    \end{tabularx}
    \\[2.5mm]
   \parbox{0.98\textwidth}{\scriptsize\raggedright
All metrics are evaluated on IAW test set, except for the 
\textbf{finite-inventory consistency rates}, which use a common 12-instance subset
because three instances lack verifiable frozen inventory references.
Compared with \textbf{Manual2Skill~\cite{tie2025manual2skill}}, \textbf{CAPER} improves the
\textbf{F1 score} by 11.25 percentage points (41.34\% vs.\ 30.09\%),
the \textbf{Finite-inventory Consistency Rate} by
41.66 percentage points (83.33\% vs.\ 41.67\%),
and the \textbf{Complete-output Finite-inventory Consistency Rate} by
25.00 percentage points (58.33\% vs.\ 33.33\%),
while reducing \textbf{Out-of-inventory Part-ID Occurrences}
from 118 to zero.
}
\end{table*}

\subsection{Results}

\subsubsection{Assembly Planning Performance}

Table~\ref{tab:overall} summarizes the primary comparison. CAPER obtains 62.99\% Structured Report Precision, 30.77\% Recall, and 41.34\% F1, achieving the highest precision and F1 among the evaluated methods. Compared with Manual2Skill~\cite{tie2025manual2skill}, CAPER increases precision from 27.08\% to 62.99\% and F1 from 30.09\% to 41.34\%, corresponding to improvements of 35.91 and 11.25 percentage points, respectively. Its Finite-inventory Consistency Rate increases from 41.67\% to 83.33\%, and its Complete-output Finite-inventory Consistency Rate increases from 33.33\% to 58.33\%, gains of 41.66 and 25.00 percentage points. Meanwhile, Out-of-inventory Part-ID Occurrences decrease from 118 to zero. These results show that CAPER improves both structured-report quality and inventory consistency over the controlled Manual2Skill  baseline. Manual2Skill retains higher recall, at 33.85\% versus 30.77\%, so CAPER's advantage is concentrated in precision, F1, and inventory consistency.

Among the three direct Qwen3-VL-4B-Instruct baselines, DPO gives the highest Structured Report F1 at 40.91\%, while SFT gives the highest precision at 62.20\%. CAPER exceeds these values by 0.43 and 0.79 percentage points, respectively. Its larger advantage appears in finite-inventory consistency: CAPER reaches $10/12$ (83.33\%), compared with $6/12$ (50.00\%), $8/12$ (66.67\%), and $6/12$ (50.00\%) for Base, SFT, and DPO. The corresponding Complete-output Finite-inventory Consistency Rates are $7/12$ (58.33\%) for CAPER, compared with $3/12$ (25.00\%), $5/12$ (41.67\%), and $4/12$ (33.33\%). CAPER also produces zero out-of-inventory Part-ID occurrences, compared with one, two, and two for these direct baselines. CAPER provides the strongest measured precision, F1, and inventory consistency in this comparison, with modest local accuracy gains over direct generation and larger gains over Manual2Skill~\cite{tie2025manual2skill}. 

\subsubsection{Analysis of the Performance Gains}

Compared with Manual2Skill~\cite{tie2025manual2skill} and Manual2Skill++~\cite{tie2026manual2skillpp}, CAPER focuses on the complementary requirements of linking each step to exact product-specific Part IDs and checking their accumulated use against finite inventory. The capability comparison in Table~I contextualizes this distinction; the quantitative evidence in Table~\ref{tab:overall} is specific to the evaluated baselines, including Manual2Skill~\cite{tie2025manual2skill}.

The semantic planner already captures much of the step-level assembly meaning from the manual. However, semantic understanding alone is insufficient for deciding which exact physical part should be used when multiple legal candidates are visually similar. PartGround addresses this gap by constraining hardware predictions to the product-specific candidate inventory and linking them with step-level manual evidence. Numeric Part IDs predicted by the VLM are not accepted as final physical identities. This evidence-guided grounding provides a mechanism for avoiding unsupported identifiers, consistent with CAPER's zero observed out-of-inventory Part-ID occurrences in Table~\ref{tab:overall}.

PlanRefine addresses a different source of error: assignments that appear plausible individually can exceed the available inventory when combined across steps. It audits the grounded assignments and removes low-confidence assignments implicated in overuse until finite-inventory consistency is restored. The shared Assembly IR connects semantic operations, exact Part IDs, quantities, and supporting evidence, allowing grounding and refinement to operate on the same assignments. This separation of responsibilities explains why CAPER's largest measured advantage over direct generation is in inventory consistency. Relative to direct structured generation and the controlled Manual2Skill procedure~\cite{tie2025manual2skill}, CAPER's contribution is the explicit combination of evidence-guided physical assignment and cross-step capacity validation. Table \ref{tab:ablation} examines these two contributions in turn.

\subsection{Ablation Study}

To address part-identity ambiguity and cross-step inventory inconsistency, our system comprises two complementary modules, PartGround and PlanRefine. We examine their effectiveness through the cumulative ablation in Table~\ref{tab:ablation}, using the configurations defined in Section~\ref{sec:caper-experimental-setup}. The two modules are analyzed separately below to distinguish their effects on local grounding and plan-level inventory consistency.

\subsubsection{PartGround}

\begin{table}[!t]
    \centering
    \caption{Ablation study of CAPER.}
    \label{tab:ablation}
    \scriptsize
    \setlength{\tabcolsep}{2.2pt}
    \renewcommand{\arraystretch}{1.13}
    \resizebox{\columnwidth}{!}{%
        \begin{tabular}{lccccc}
            \toprule
            Method
            & \shortstack{F1 Score\\(\%) $\uparrow$}
            & \shortstack{Step Report\\Accuracy (\%) $\uparrow$}
            & \shortstack{Part-ID F1\\Score (\%) $\uparrow$}
            & \shortstack{Hardware Exact\\Accuracy (\%) $\uparrow$}
            & \shortstack{Finite-inventory\\Consistency Rate (\%) $\uparrow$} \\
            \midrule
            \rowcolor{green!15}
            \textbf{CAPER (Ours)}
            & \textbf{41.34}
            & \textbf{23.23}
            & \textbf{55.77}
            & \textbf{42.42}
            & \textbf{83.33} \\
            w/o PlanRefine
            & 41.18 & \textbf{23.23} & 55.24 & 41.75 & 58.33 \\
            \shortstack[l]{w/o PartGround\\and PlanRefine}
            & 40.77 & 22.90 & 54.31 & 41.08 & 58.33 \\
            \bottomrule
        \end{tabular}%
    }
    \vspace{1mm}
\end{table}

Table\ref{tab:ablation} shows that adding PartGround to the semantic planner increases Part-ID F1 from 54.31\% to 55.24\% and Hardware Exact Accuracy from 41.08\% to 41.75\%, gains of 0.93 and 0.67 percentage points, respectively. Structured Report F1 increases from 40.77\% to 41.18\%, and Step Report Accuracy increases from 22.90\% to 23.23\%. Finite-inventory Consistency Rate remains unchanged at 58.33\%. These results indicate that explicit physical grounding provides modest improvements in local identity and hardware prediction, while grounding alone does not improve the measured cross-step inventory consistency.

This pattern is consistent with the design of PartGround in Section~\ref{subsec:partground}. For each step, the module forms an inventory-constrained proposal from the localized manual evidence and candidate references, while also searching for explicit hardware callouts and quantity annotations. Agreement between these evidence sources provides stronger support; forward-only evidence is retained with lower support, and disagreement is not silently overwritten. These mechanisms connect generic hardware mentions to specific physical candidates while preserving evidence for review. This is consistent with the smaller gain in overall Structured Report F1, which also reflects operation facts supplied by the semantic planner.

These results show PartGround's contribution to exact physical assignment and hardware-level prediction. Its distinctive role is to transform free-form hardware prediction into evidence-guided, product-specific Part IDs and quantity assignments within the Assembly IR. This extends structured manual interpretation, as represented by Manual2Skill~\cite{tie2025manual2skill}, with an explicit grounding stage, although discrimination among visually similar legal candidates remains difficult.

\subsubsection{PlanRefine}

Adding PlanRefine to the PartGround configuration increases Finite-inventory Consistency Rate from $7/12$ (58.33\%) to $10/12$ (83.33\%), a gain of 25.00 percentage points in Table\ref{tab:ablation}. In comparison, Structured Report F1 changes from 41.18\% to 41.34\%, and Part-ID F1 increases from 55.24\% to 55.77\%. Hardware Exact Accuracy rises from 41.75\% to 42.42\%, while Step Report Accuracy remains unchanged at 23.23\%. Thus, the primary demonstrated effect of PlanRefine is improved finite-inventory consistency, accompanied by small changes in local prediction quality.

This outcome follows from the capacity-repair mechanism in Section~\ref{subsec:planrefine}. PlanRefine aggregates each Part ID's predicted consumption across the complete assembly sequence and compares it with the available quantity. When consumption exceeds inventory, it removes assignments in increasing order of grounding confidence, breaking ties by removing later-step assignments first, and stops once capacity is restored. It then re-audits the repaired plan against the same inventory. PlanRefine does not infer replacement Part IDs or search alternative candidates, so its main effect is to resolve cross-step over-allocation rather than improve local identity discrimination.

PlanRefine substantially improves the measured inventory consistency while preserving the aggregate level of local prediction accuracy. Its contribution is an explicit, confidence-guided capacity repair applied to grounded assignments, complementing PartGround's evidence-guided identity selection. Together, the two modules extend step-level structured prediction with both physical grounding and finite-inventory validation, addressing reliability requirements that are not ensured by the controlled direct-generation and Manual2Skill baselines~\cite{tie2025manual2skill}.

\subsection{Robotic Assembly Experiments}

To investigate the downstream impact of part grounding on robotic execution, we conducted furniture assembly experiments in NVIDIA IsaacSim~\cite{nvidia_IsaacSim} using a Franka Emika Panda manipulator
\cite{haddadin2022franka}.The simulation experiments were run on a server equipped with four NVIDIA GeForce RTX 5090 GPUs and covered five furniture instances. These experiments examine whether identifying the exact parts required by an instruction step improves the robot's ability to select the intended components and carry out the corresponding assembly operation.

We compared assembly execution with and without explicit Part-ID grounding. Without grounding, we observed incorrect part selection and failures to initiate required grasps when the intended component was ambiguous. With CAPER, the execution system received explicit Part-ID and quantity assignments through the Assembly IR. The success rate in simulation improved significantly.

Across the five evaluated furniture instances, execution with CAPER exhibited improved part-selection accuracy and more reliable assembly-step completion compared with execution without explicit Part-ID grounding. The observed improvement illustrates the practical value of resolving part identities before robot execution: selecting the correct component is a prerequisite for performing the intended assembly operation. These findings support the practical value of CAPER's part grounding for improving part selection and assembly execution reliability in the evaluated simulation tasks.

\begin{figure}[!t]
    \centering
    \includegraphics[width=\columnwidth]{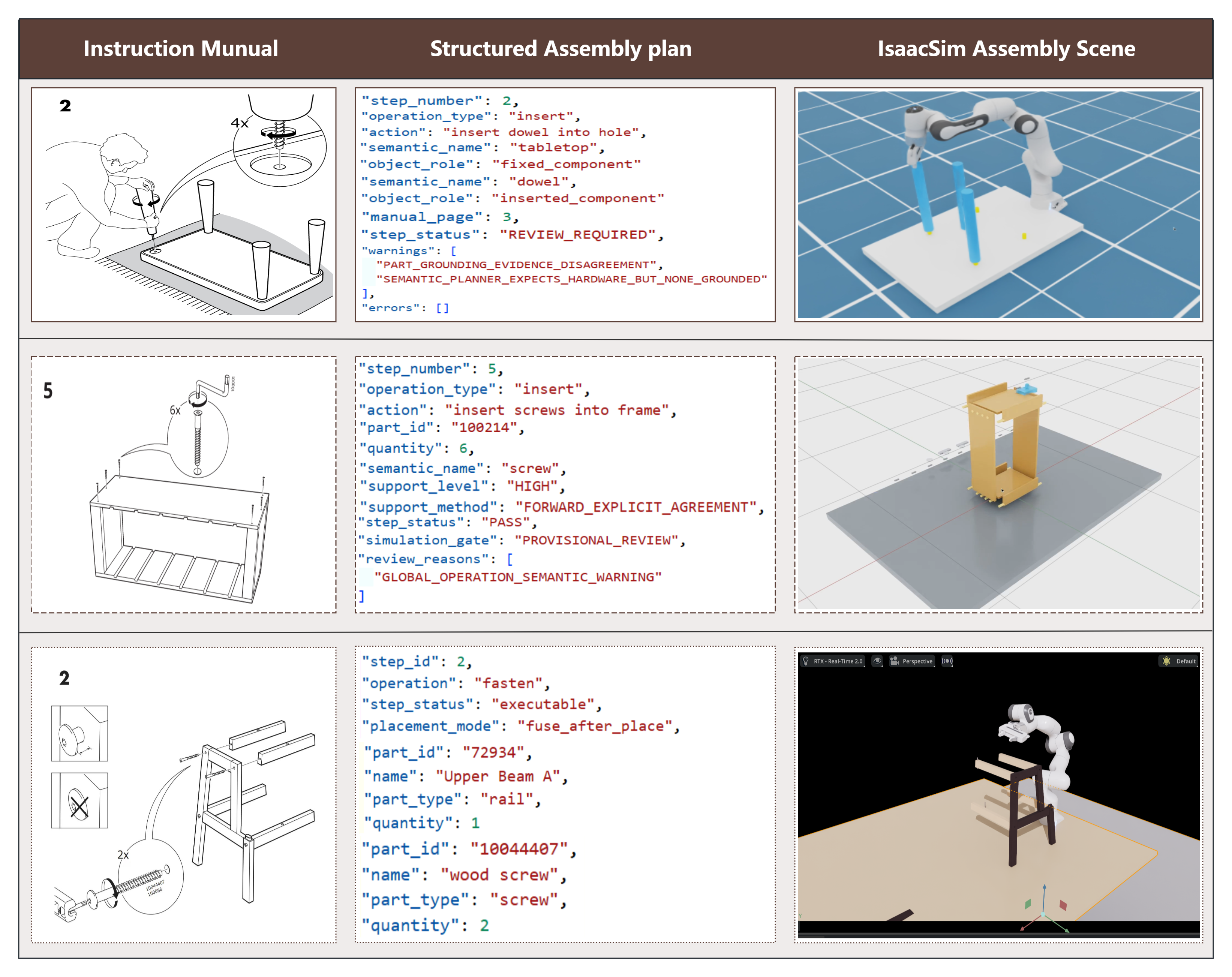}
    \caption[Representative furniture assembly examples.]{%
        \textbf{Representative furniture assembly examples.}
    }
    \label{fig:robot_assembly}
    \par\vspace{2.5mm}
    \parbox{0.98\columnwidth}{\scriptsize\raggedright
        From top to bottom, the rows show MAMMUT (Step 2),
        TROFAST (Step 5), and BEKV{\"A}M (Step 2).
        \par
        Left: assembly illustrations from the instruction manuals.
        Middle: corresponding structured assembly plan excerpts,
        including assembly operations, part information, hardware
        quantities, and execution or review status.
        Right: corresponding assembly scenes in NVIDIA IsaacSim.
    }
\end{figure}

Fig.~\ref{fig:robot_assembly} shows three representative steps (MAMMUT Step 2, TROFAST Step 5, and BEKV{\"A}M Step 2). From left to right, each row aligns the instruction-manual illustration, the structured Assembly IR excerpt (operation, Part IDs, quantities, and PASS or REVIEW REQUIRED status), and the corresponding NVIDIA IsaacSim scene. These views make the mapping from manual evidence to physical assignment and simulated execution inspectable.

    \section{Discussion and Limitations}
    The current IAW test set [8] results show that CAPER improves both local prediction quality and plan-level reliability. CAPER achieves a Structured Report F1 of 41.34\%, compared with 40.91\% for the strongest direct Qwen3-VL-4B-Instruct baseline, while reaching an 83.33\% Finite-inventory Consistency Rate and a 58.33\% Complete-output Finite-inventory Consistency Rate with zero observed inventory violations. These gains reflect the complementary roles of the two modules: PartGround improves evidence-guided physical grounding, whereas PlanRefine strengthens cross-step finite-inventory consistency through minimal refinement of conflicting assignments. 
    
   Furthermore, the full CAPER system incurs considerable computational overhead originating from layout parsing, multi-region visual‑evidence matching and VLM inference, rather than an exact combinatorial inventory‑assignment solver.
    
    \section{Conclusion}
    We presented CAPER, a manual-guided robotic assembly framework for Part-ID grounding and finite-inventory consistency. The VLM is used only as a semantic planner. PartGround converts step-level manual evidence into inventory-constrained Part IDs and quantity assignments, while PlanRefine audits those assignments against finite inventory without re-inferring identities. Both write to a shared Assembly IR, which retains review states when the evidence is insufficient.

    On the IAW test set~\cite{zhang2023iaw}, the gain is generally above peer level. CAPER reaches a Finite-inventory Consistency Rate of 83.33\% and produces zero out-of-inventory Part-ID occurrences, while Structured Report F1 remains above the strongest direct Qwen3-VL-4B-Instruct baseline. PartGround turns free-form hardware mentions into inventory-legal Part-ID assignments; PlanRefine then raises inventory consistency from 58.33\% to 83.33\% and removes all observed overuse pairs ($9\rightarrow 0$) and excess units ($51\rightarrow 0$).

    The Assembly IR is the interface to execution. It aligns manual instructions with exact Part IDs, quantities, and a PASS or REVIEW REQUIRED state for the corresponding IsaacSim scene. Thus,CAPER turns VLM-generated assembly knowledge into inventory-consistent physical targets for downstream perception, grasping, and execution.

    \section*{Acknowledgment}
    Generative AI (OpenAI ChatGPT) was used for initial drafting, language organization, and schematic figure ideation. All technical claims, experimental results, code, figures, and references remain the authors' responsibility.
    
\bibliographystyle{IEEEtran}
\bibliography{references}

@inproceedings{tie2025manual2skill,
  author    = {Chenrui Tie and Shengxiang Sun and Jinxuan Zhu and Yiwei Liu and Jingxiang Guo and Yue Hu and Haonan Chen and Junting Chen and Ruihai Wu and Lin Shao},
  title     = {Manual2Skill: Learning to Read Manuals and Acquire Robotic Skills for Furniture Assembly Using Vision-Language Models},
  booktitle = {Proceedings of Robotics: Science and Systems},
  year      = {2025},
  address   = {Los Angeles, CA, USA},
  month     = {June},
  doi       = {10.15607/RSS.2025.XXI.150}
}

@inproceedings{tie2026manual2skillpp,
  author    = {Chenrui Tie and Shengxiang Sun and Yudi Lin and Yanbo Wang and Zhongrui Li and Zhouhan Zhong and Jinxuan Zhu and Yiman Pang and Haonan Chen and Junting Chen and Ruihai Wu and Lin Shao},
  title     = {Manual2Skill++: Connector-Aware General Robotic Assembly from Instruction Manuals via Vision-Language Models},
  booktitle = {2026 IEEE International Conference on Robotics and Automation (ICRA)},
  year      = {2026},
  note      = {arXiv:2510.16344}
}

@inproceedings{yang2025vlmtamp,
  author    = {Zhutian Yang and Caelan Garrett and Dieter Fox and Tom{\'a}s Lozano-P{\'e}rez and Leslie Pack Kaelbling},
  title     = {Guiding Long-Horizon Task and Motion Planning with Vision Language Models},
  booktitle = {2025 IEEE International Conference on Robotics and Automation (ICRA)},
  year      = {2025},
  pages     = {16847--16853},
  doi       = {10.1109/ICRA55743.2025.11128705}
}

@inproceedings{ren2023knowno,
  author    = {Allen Z. Ren and Anushri Dixit and Alexandra Bodrova and Sumeet Singh and Stephen Tu and Noah Brown and Peng Xu and Leila Takayama and Fei Xia and Jake Varley and Zhenjia Xu and Dorsa Sadigh and Andy Zeng and Anirudha Majumdar},
  title     = {Robots That Ask For Help: Uncertainty Alignment for Large Language Model Planners},
  booktitle = {Proceedings of the 7th Conference on Robot Learning},
  series    = {Proceedings of Machine Learning Research},
  volume    = {229},
  pages     = {661--682},
  year      = {2023},
  publisher = {PMLR}
}

@inproceedings{ao2025btplanner,
  author    = {Jicong Ao and Fan Wu and Yansong Wu and Abdalla Swikir and Sami Haddadin},
  title     = {LLM-as-BT-Planner: Leveraging LLMs for Behavior Tree Generation in Robot Task Planning},
  booktitle = {2025 IEEE International Conference on Robotics and Automation (ICRA)},
  year      = {2025},
  pages     = {1233--1239},
  doi       = {10.1109/ICRA55743.2025.11128454}
}

@inproceedings{singh2023progprompt,
  author    = {Ishika Singh and Valts Blukis and Arsalan Mousavian and Ankit Goyal and Danfei Xu and Jonathan Tremblay and Dieter Fox and Jesse Thomason and Animesh Garg},
  title     = {ProgPrompt: Generating Situated Robot Task Plans using Large Language Models},
  booktitle = {2023 IEEE International Conference on Robotics and Automation (ICRA)},
  year      = {2023},
  pages     = {11523--11530},
  doi       = {10.1109/ICRA48891.2023.10161317}
}

@inproceedings{ichter2022saycan,
  author    = {Brian Ichter and Anthony Brohan and Yevgen Chebotar and Chelsea Finn and Karol Hausman and Alexander Herzog and Daniel Ho and Julian Ibarz and Alex Irpan and Eric Jang and Ryan Julian and Dmitry Kalashnikov and Sergey Levine and Yao Lu and Carolina Parada and Kanishka Rao and Pierre Sermanet and Alexander Toshev and Vincent Vanhoucke and Fei Xia and Ted Xiao and Peng Xu and Mengyuan Yan and Noah Brown and Michael Ahn and Omar Cortes and Nicolas Sievers and Clayton Tan and Sichun Xu and Diego Reyes and Jarek Rettinghouse and Jornell Quiambao and Peter Pastor and Linda Luu and Kuang-Huei Lee and Yuheng Kuang and Sally Jesmonth and Nikhil J. Joshi and Kyle Jeffrey and Rosario Jauregui Ruano and Jasmine Hsu and Keerthana Gopalakrishnan and Byron David and Andy Zeng and Chuyuan Kelly Fu},
  title     = {Do As I Can, Not As I Say: Grounding Language in Robotic Affordances},
  booktitle = {Proceedings of the 6th Conference on Robot Learning},
  series    = {Proceedings of Machine Learning Research},
  volume    = {205},
  pages     = {287--318},
  year      = {2023},
  publisher = {PMLR}
}

@inproceedings{huang2023inner,
  author    = {Wenlong Huang and Fei Xia and Ted Xiao and Harris Chan and Jacky Liang and Pete Florence and Andy Zeng and Jonathan Tompson and Igor Mordatch and Yevgen Chebotar and Pierre Sermanet and Tomas Jackson and Noah Brown and Linda Luu and Sergey Levine and Karol Hausman and Brian Ichter},
  title     = {Inner Monologue: Embodied Reasoning through Planning with Language Models},
  booktitle = {Proceedings of the 6th Conference on Robot Learning},
  series    = {Proceedings of Machine Learning Research},
  volume    = {205},
  pages     = {1769--1782},
  year      = {2023},
  publisher = {PMLR}
}

@inproceedings{shao2026symskill,
  author    = {Yifei Shao and Yuchen Zheng and Sunan Sun and Pratik Chaudhari and Vijay Kumar and Nadia Figueroa},
  title     = {SymSkill: Symbol and Skill Co-Invention for Data-Efficient and Reactive Long-Horizon Manipulation},
  booktitle = {2026 IEEE International Conference on Robotics and Automation (ICRA)},
  year      = {2026},
  note      = {arXiv:2510.01661}
}

@article{bai2025qwen3vl,
  author  = {Shuai Bai and Yuxuan Cai and Ruizhe Chen and Keqin Chen and Xionghui Chen and Zesen Cheng and Lianghao Deng and Wei Ding and Chang Gao and Chunjiang Ge and Wenbin Ge and Zhifang Guo and Qidong Huang and Jie Huang and Fei Huang and Binyuan Hui and Shutong Jiang and Zhaohai Li and Mingsheng Li and Mei Li and Kaixin Li and Zicheng Lin and Junyang Lin and Xuejing Liu and Jiawei Liu and Chenglong Liu and Yang Liu and Dayiheng Liu and Shixuan Liu and Dunjie Lu and Ruilin Luo and Chenxu Lv and Rui Men and Lingchen Meng and Xuancheng Ren and Xingzhang Ren and Sibo Song and Yuchong Sun and Jun Tang and Jianhong Tu and Jianqiang Wan and Peng Wang and Pengfei Wang and Qiuyue Wang and Yuxuan Wang and Tianbao Xie and Yiheng Xu and Haiyang Xu and Jin Xu and Zhibo Yang and Mingkun Yang and Jianxin Yang and An Yang and Bowen Yu and Fei Zhang and Hang Zhang and Xi Zhang and Bo Zheng and Humen Zhong and Jingren Zhou and Fan Zhou and Jing Zhou and Yuanzhi Zhu and Ke Zhu},
  title   = {Qwen3-VL Technical Report},
  journal = {arXiv preprint arXiv:2511.21631},
  year    = {2025}
}

@inproceedings{rafailov2023dpo,
  author    = {Rafael Rafailov and Archit Sharma and Eric Mitchell and Christopher D. Manning and Stefano Ermon and Chelsea Finn},
  title     = {Direct Preference Optimization: Your Language Model Is Secretly a Reward Model},
  booktitle = {Advances in Neural Information Processing Systems},
  volume    = {36},
  year      = {2023}
}

@inproceedings{zhang2023iaw,
  author    = {Jiahao Zhang and Anoop Cherian and Yanbin Liu and Yizhak Ben-Shabat and Cristian Rodriguez and Stephen Gould},
  title     = {Aligning Step-by-Step Instructional Diagrams to Video Demonstrations},
  booktitle = {Proceedings of the IEEE/CVF Conference on Computer Vision and Pattern Recognition (CVPR)},
  year      = {2023},
  pages     = {2483--2492},
  doi       = {10.1109/CVPR52729.2023.00245}
}

@inproceedings{sera2021graphicalmanual,
  author    = {Issei Sera and Natsuki Yamanobe and Ixchel G. Ramirez-Alpizar and Zhenting Wang and Weiwei Wan and Kensuke Harada},
  title     = {Assembly Planning by Recognizing a Graphical Instruction Manual},
  booktitle = {2021 IEEE/RSJ International Conference on Intelligent Robots and Systems (IROS)},
  year      = {2021},
  pages     = {3138--3145},
  doi       = {10.1109/IROS51168.2021.9636041}
}

@misc{nvidia_isaacsim,
  author       = {{NVIDIA}},
  title        = {{Isaac Sim}},
  howpublished = {\url{https://github.com/isaac-sim/IsaacSim}},
  note         = {Version 6.0.1, accessed Sep. 16, 2026}
}

@article{haddadin2022franka,
  author  = {Haddadin, Sami and Parusel, Sven and Johannsmeier, Lars
             and Golz, Saskia and Gabl, Simon and Walch, Florian
             and Sabaghian, Mohamadreza and J{\"a}hne, Christoph
             and Hausperger, Lukas and Haddadin, Simon},
  title   = {The Franka Emika Robot: A Reference Platform for Robotics Research and Education},
  journal = {IEEE Robotics \& Automation Magazine},
  volume  = {29},
  number  = {2},
  pages   = {46--64},
  year    = {2022},
  doi     = {10.1109/MRA.2021.3138382}
}
\end{document}